\documentclass[a4paper, 10 pt, conference]{cras} 
\usepackage[utf8]{inputenc}
\IEEEoverridecommandlockouts               
\usepackage{mathtools}
\usepackage{geometry}
\usepackage{newtxmath,newtxtext}
\usepackage{authblk}
\usepackage{pgfplots}
\usepackage{todonotes}
\usepackage{subfig}
\usepackage{graphicx}
\usepackage{color}
\usepackage{float}
\usepackage{soul}
\usepackage{verbatim}

\pgfplotsset{compat=newest} 
\pgfplotsset{plot coordinates/math parser=false} 
 
\title{\Large \bf A Fabric Soft Robotic Exoskeleton with Novel Elastic Band Integrated Actuators for Hand Rehabilitation
} 

\author{\large Cem Suulker, Sophie Skach, Kaspar Althoefer \\ \\ \small \textit{Centre for Advanced Robotics}\\ School of Engineering and Materials Science \\Queen Mary University of London, United Kingdom}

\begin{document}

\maketitle
\thispagestyle{empty}
\pagestyle{empty}

\section*{INTRODUCTION}
Common disabilities like stroke and spinal cord injuries may cause loss of motor function in hands \cite{takahashi2008robot}. They can be treated with robot assisted rehabilitation techniques, like continuously opening and closing the hand with help of a robot, in a cheaper, and less time consuming manner than traditional methods \cite{wolf2006effect}. Hand exoskeletons are developed to assist rehabilitation \cite{kawasaki2007development}, but their bulky nature brings with it certain challenges. As soft robots use elastomeric and fabric elements rather than heavy links\cite{9762074}, and operate with pneumatic, hydraulic or tendon based rather than traditional rotary or linear motors, soft hand exoskeletons are deemed a better option in relation to rehabilitation \cite{radder2018feasibility}.

\begin{figure}[h]
  \centering
  \includegraphics[width=0.6\linewidth]{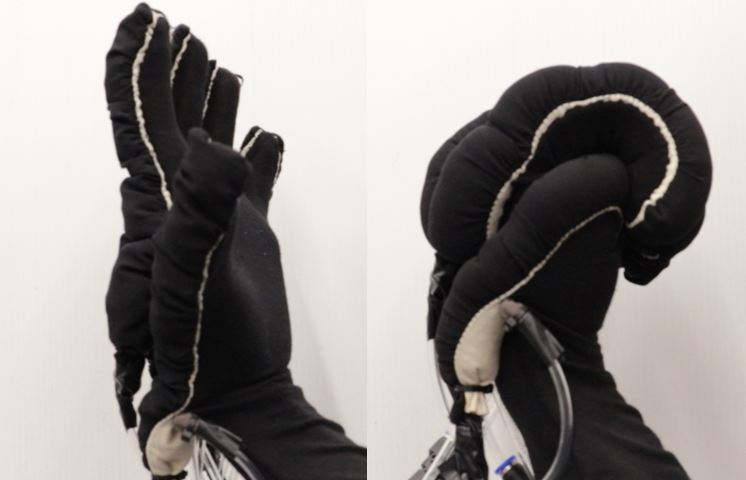}
  \caption{Our novel fabric based soft hand exoskeleton robot, worn relaxed (left) and actuated (right).}
  \label{fig1}
\end{figure}

A variety of soft actuator types can be used to achieve hand rehabilitation tasks. Silicone \cite{polygerinos2015soft2}, fabric \cite{yap2017fully} and tendon \cite{in2015exo} based actuators are the most common ones. In this work we present a soft hand exoskeleton introducing elastic band integrated fabric based actuators (Fig. \ref{fig1}). The glove helps the user perform hand flexion and extension motions, making it usable for rehabilitation and assistive purposes. Furthermore, thanks to expert textile manufacturing techniques and the exploiting of soft textile structures, it enhances wearability comfort and is closer than previous incarnations to a clothing accessory.

\section*{MATERIALS AND METHODS}
Our wearable robot is developed following iterative prototyping and experimenting with various textile materials. To optimise the design, we have tested a wide selection of fabric types and manufacturing techniques that show potential synergy for a soft exoskeleton finger. Due to required stretch capabilities, we focused on structures with a high elasticity as a key characteristic. Moreover, we chose materials that accommodate ubiquitous design approaches to rehabilitation equipment and a better user-acceptance: soft and smooth fabrics similar to our own clothing.  

Among the other key features of fabric actuators for soft hand exoskeletons are a large curvature and constant force across the surface of the actuator.
This can be achieved by creating an imbalance between the top and bottom layer of the fabrics. The greater the imbalance, the better the flexion as well as force capability.

\begin{figure}[h]
  \centering
  \includegraphics[width=0.75\linewidth]{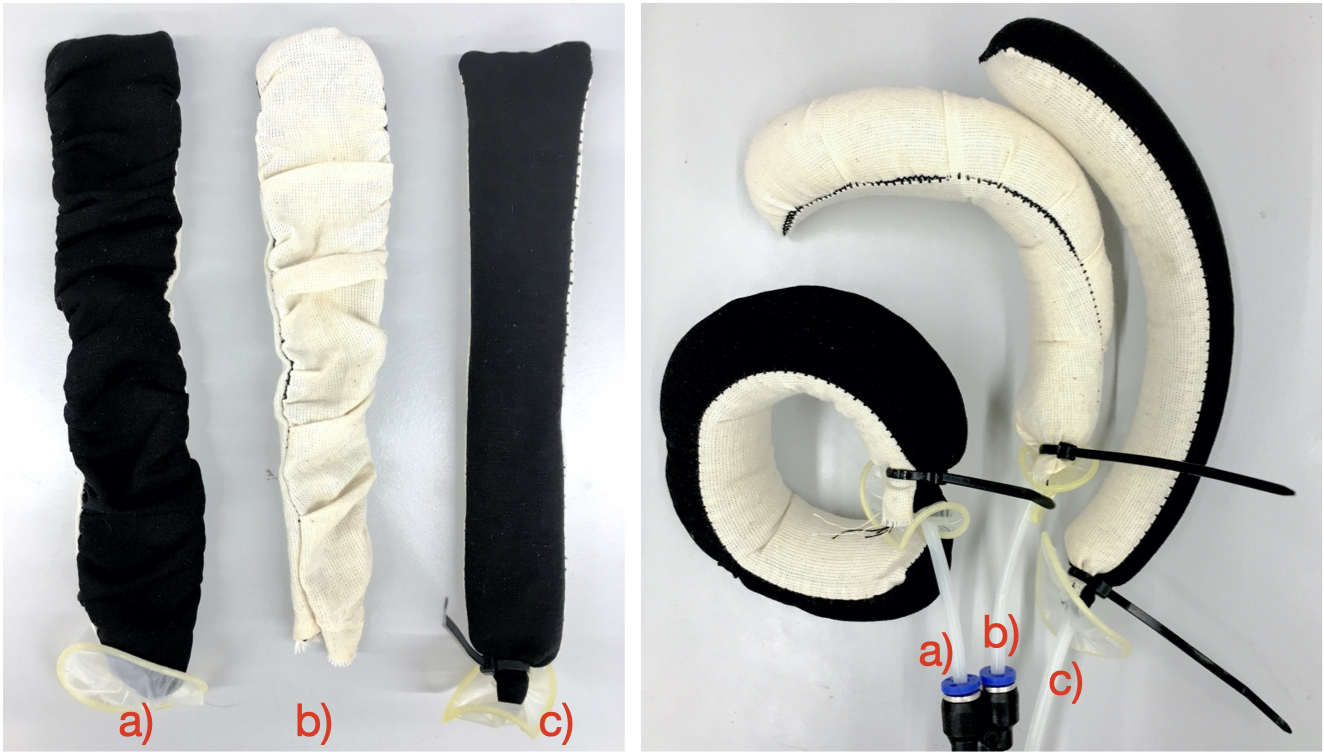}
  \caption{Best performing designs non actuated (left) and actuated (right): a)woven stretch fabric, elastic band ; b) non-elastic woven, elastic band; c) woven stretch, no band.}
  \label{fig3}
\end{figure}

After preliminary sampling, we selected 3 actuators, shown in Fig. \ref{fig3}, that combined the most successful actuator characteristics: a woven stretch fabric (Fig. \ref{fig3} c); a non-stretch woven cotton fabric with an added elastic band along the sides (Fig. \ref{fig3} b); and the combination of the two, a woven stretch fabric with elastic band (Fig. \ref{fig3} a). To actuate them, a latex bladder is inserted to guarantee airtightness.

\section*{RESULTS}
We ran tests to measure selected actuator's flexion angle and force capabilities. Flexion angle is a critical parameter for assistive gloves, it is imperative that each finger is unrestricted in relation to its maximum angle \cite{takahashi2008robot}. Inflatable actuators therefore need large bending capabilities. Equally important is its force capability. For achieving activities of daily living an acceptable force capability is around 10-15 N \cite{takahashi2008robot}.

\begin{figure}[H]
\centering
  \includegraphics[width=1\linewidth]{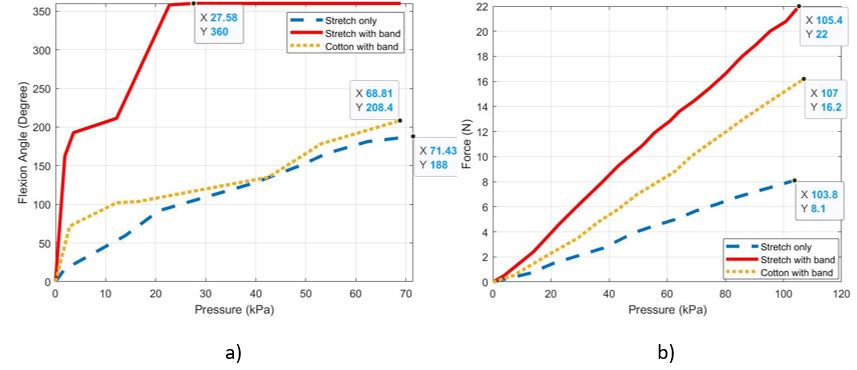}
  \caption{Flexion angle versus pressure (a) and force versus pressure (b) plots for three selected actuators.}
  \label{figf2}
\end{figure}

Figure \ref{figf2}(a) shows that bending angle capability of the 'stretch with band' actuator is superior to the others. Just 30 kPa pressure is enough for it to bend a full cycle of 360 degrees. 'Stretch only' and 'cotton with band' actuators can bend approximately 190 and 210 degrees respectively under 70 kPa pressure. While these values are also acceptable from a rehabilitation perspective, the 'stretch with band' actuator is clearly a better option.

Figure \ref{figf2}(b) shows the 'stretch with band' actuator can apply 22 N force at about 100 kPa pressure. This is outperforming other fabric or silicone based soft hand exoskeleton bending actuators. The 'stretch only' actuator can go up to 8 N and the 'cotton with elastic band' up to 18 N.

To unleash the maximum potential of the actuators, they need to be inflated to 100 kPa. However to prevent the latex bladder to fail (between 110 and 120 kPa) running them at 100 kPa for a long period might reduce the lifetime of the bladder. Operating them at 70 kPa might represent a sensible compromise, but that pressure  would not be enough for the 'stretch only' actuator to carry out activities of daily living. The 'stretch with band' and 'cotton with elastic band', however, would be able to do so \cite{takahashi2008robot}.

These tests confirm that, the actuator using stretch fabric with integrated elastic bands shows excellent performance both in relation to bending angle potential and force capability. The findings of the tests can now be used to lead the design of a new exoskeleton prototype.

\subsection*{The Prototype}

The fabric based soft hand exoskeleton is manufactured using our 'stretch with band' actuator (Figure \ref{fig1}) which was found to be the best candidate in light of test results. The base glove, which the actuators are mounted on, is made from four-way stretch fabric: a black viscose jersey knit to guarantee soft feel on the skin \cite{9844801}.

The five inflatable fingers were sewn together with an overlock machine and hand-stitched onto the glove, reaching from the fingertip down to the centre of the back of the palm. For the glove to be able to assist with the extension motion of the hand as well, another simple two layer fabric actuator is hand stitched between the base glove and the bending actuators.

\begin{figure}[h]
  \centering
  \includegraphics[width=0.7\linewidth]{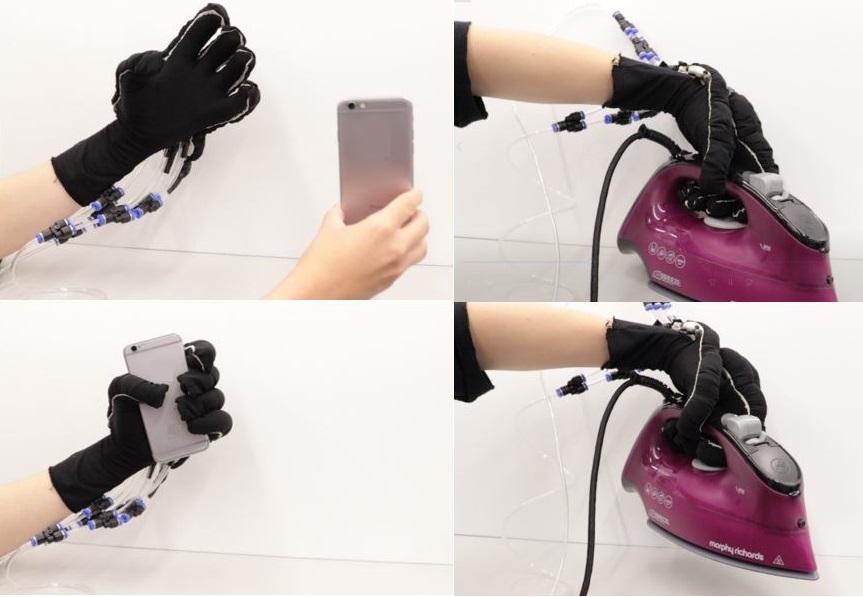}
  \caption{The final hand exoskeleton glove ``in action" grasping and releasing a phone and an iron.}
  \label{figg2}
\end{figure}

In Figure \ref{figg2} the exoskeleton glove can be seen supporting the hand while grasping objects. 
 
\section*{CONCLUSION AND DISCUSSION}


In this work, we present a novel actuation method for a fabric-based soft robotic exoskeleton glove. Integrating elastic band to the stretch fabric based actuator enhancing the capability of the exoskeleton, making it able to bend high angles in low pressure values, and boosting its force exertion capability to the fingers.
Furthermore, our ’stretch with band’ actuator can produce higher forces than any other available silicone or fabric based alternative. This improves grasping ability for the user that is a critical element in the rehabilitation process.

Assistive wearable devices should be designed with consideration given to the needs of the physically vulnerable. Being lightweight, a characteristic inherent in fabric-based devices, is therefore clearly a potential advantage textiles bear. We are accustomed to feel textile on our skin, which makes this material promising with regards to rehabilitation. Last but not least, this makes the robot also more acceptable to users.


\bibliographystyle{IEEEtran}
\bibliography{CRAS}

\end{document}